\documentclass{article}

\usepackage{microtype}
\usepackage{graphicx}
\usepackage{subcaption}
\usepackage{booktabs}

\usepackage{hyperref}

\usepackage[preprint]{style}

\usepackage{amsmath}
\usepackage{amssymb}
\usepackage{mathtools}
\usepackage{amsthm}
\usepackage{overpic}
\usepackage{multirow}
\usepackage{multicol}
\usepackage{tikz}

\usepackage[capitalize,noabbrev]{cleveref}

\theoremstyle{plain}

\theoremstyle{definition}

\theoremstyle{remark}

\usepackage[textsize=tiny]{todonotes}

\usepackage{cuted}   
\usepackage{capt-of} 
\usepackage{multirow}

\usepackage{xspace}
\makeatletter
\DeclareRobustCommand\onedot{\futurelet\@let@token\@onedot}
\def\@onedot{\ifx\@let@token.\else.\null\fi\xspace}
\makeatother

\def\eg{\emph{e.g}\onedot}

\definecolor{mypink}{RGB}{255,200,190} 

\usepackage{colortbl}
\definecolor{myblue}{RGB}{203,203,254}

\usepackage[dvipsnames, table]{xcolor}
\usepackage{enumitem}
\usepackage[many]{tcolorbox} 
\tcbuselibrary{listings, breakable}
\newtcolorbox{promptBox}[1]{
    title=\textbf{#1},
    breakable,
    fonttitle=\bfseries,
    boxrule = 1pt,
    toprule = 3pt, 
    colframe = RoyalBlue,
    enhanced,
    rounded corners,
    arc = 2pt, 
    top=1mm,bottom=1mm,left=1mm,right=1mm  
}

\titlerunning{LayerT2V: A Unified Multi-Layer Video Generation Framework}

\begin{document}

\twocolumn[
  \titleline{LayerT2V: A Unified Multi-Layer Video Generation Framework}

  \ssetsymbol{equal}{*}
  \ssetsymbol{cor}{$\dagger$}
  \ssetsymbol{sii}{5}
  \begin{authorlist}
    \authorline{Guangzhao Li}{sjtu,sii,equal}
    \authorline{Kangrui Cen}{2,3,equal}
    \authorline{Baixuan Zhao}{sjtu}
    \authorline{Yi Xin}{4,5}
    \authorline{Siqi Luo}{sjtu} 
    \authorline{Guangtao Zhai}{sjtu}
    \authorline{Lei Zhang}{2,3}
    \authorline{Xiaohong Liu}{sjtu,sii,cor}
  \end{authorlist}
 \begin{center}
     $^{1}$Shanghai Jiao Tong University \quad  $^{2}$Hong Kong Polytechnic University \quad $^{3}$OPPO Research Institute  \\  $^{4}$Nanjing University  \quad  $^{5}$Shanghai Innovation Institute \\ 
 \end{center}
 \begin{center}
    \url{https://layert2v.github.io}
  \end{center}

  \vskip 0.3in
]

{
  \renewcommand{\thefootnote}%
    {\fnsymbol{footnote}}
 \footnotetext[1]{Equal contribution; $^{\dagger}$Corresponding author.
  }
}

\begin{abstract}
Text-to-video generation has advanced rapidly, but existing methods typically output only the final composited video and lack editable layered representations, limiting their use in professional workflows. We propose \textbf{LayerT2V}, a unified multi-layer video generation framework that produces multiple semantically consistent outputs in a single inference pass: the full video, an independent background layer, and multiple foreground RGB layers with corresponding alpha mattes. Our key insight is that recent video generation backbones use high compression in both time and space, enabling us to serialize multiple layer representations along the temporal dimension and jointly model them on a shared generation trajectory. This turns cross-layer consistency into an intrinsic objective, improving semantic alignment and temporal coherence. To mitigate layer ambiguity and conditional leakage, we augment a shared DiT backbone with LayerAdaLN and layer-aware cross-attention modulation. LayerT2V is trained in three stages: alpha mask VAE adaptation, joint multi-layer learning, and multi-foreground extension. We also introduce \textbf{VidLayer}, the first large-scale dataset for multi-layer video generation. Extensive experiments demonstrate that LayerT2V substantially outperforms prior methods in visual fidelity, temporal consistency, and cross-layer coherence.
\end{abstract}
\section{Introduction}
\label{sec:intro}

In recent years, diffusion-based text-to-video (T2V) generation has made remarkable progress. Models such as Sora V2~\cite{openai2024sora}, Wan~\cite{wan2025wan}, and HunyuanVideo~\cite{kong2024hunyuanvideo} can synthesize high-quality videos with complex motion and rich visual details from text prompts. However, the prevailing paradigm treats a video as a single, complete result rather than a collection of semantically separable layers, limiting its applicability in professional production scenarios that require precise control and local edits.

In real-world production pipelines, videos are rarely edited as a single block. Instead, creators rely on layered representations where foreground, background, and alpha matte are handled separately and then composed together, enabling flexible edits such as replacing backgrounds, refining subjects, or applying localized effects. Existing T2V models~\cite{yang2024cogvideox, wan2025wan, kong2024hunyuanvideo, openai2024sora} output only the final composited video without explicit layer decomposition (\eg, foreground, background, and alpha), leaving subsequent compositing and localized editing without a direct manipulation space.

Prior studies have explored layered generation under restricted settings, mostly focusing on producing a single RGBA foreground for images~\cite{zhang2024transparent, quattrini2024alfie} or videos~\cite{dong2025video}, while lacking explicit background modeling and cross-layer consistency constraints. LayerFlow~\cite{ji2025layerflow} investigates multi-layer joint generation, but due to limited high-quality data and the absence of explicit hierarchical interaction modeling, its results still suffer from limited stability and cross-layer coherence. Developing a unified architecture that produces high-quality, cross-layer consistent, and editable multi-layer videos remains a central challenge.

To address this, we propose \textbf{LayerT2V}, a unified framework that generates multi-layer videos in a single denoising process. Given text prompts, LayerT2V simultaneously produces multiple semantically consistent outputs: the composited video, an independent background layer, and one or more foreground RGB layers with their corresponding alpha mattes. Our key insight is that recent video diffusion models use very high compression in both time and space~\cite{blattmann2023align, wan2025wan}, making it feasible to jointly model multiple layers along a shared denoising trajectory. This turns cross-layer consistency from an external post-processing constraint into an intrinsic generation objective.

However, directly extending existing architectures to the multi-layer setting introduces new challenges: unified generation can cause discontinuities near inter-layer boundaries, and different layers exhibit substantially different statistics, such as alpha mattes that are sparse and near-binary, fundamentally different from content-rich RGB layers. To address these issues, we introduce \textbf{LayerAdaLN} for layer-specific feature modulation, and \textbf{LayeredCrossAttention} to control layer-wise interactions with text conditions. Additionally, we finetune the Wan VAE~\cite{wan2025wan} for alpha mask processing to improve matte reconstruction and stabilize multi-layer generation.

High-quality training data is a major bottleneck for multi-layer video generation. While recent large-scale video-text datasets have fueled T2V training~\cite{Bain21,chen2024panda070m0,wang2023internvid0, DBLP:conf/iclr/NanXZFYCL0T25}, they do not provide layer-aligned supervision. We develop an automated pipeline for multi-layer data construction and cleaning, and build \textbf{VidLayer}, the first large-scale multi-layer video dataset containing approximately 4M frames. Each sample includes the full video, background layer, foreground layer, alpha matte, and fine-grained layer-level text descriptions. VidLayer fills the gap of high-quality multi-layer video data and provides a scalable, controllable, and evaluable foundation for multi-layer video generation.

In summary, our main contributions are:
\begin{itemize}
    \item We propose \textbf{LayerT2V}, a unified framework that simultaneously produces multiple semantically consistent layer representations in a single inference pass.
    \item We construct \textbf{VidLayer}, the first large-scale multi-layer video dataset, providing a scalable and evaluable data foundation for multi-layer video generation.
    \item We introduce \textbf{LayerAdaLN} and \textbf{LayeredCrossAttention} to enable explicit layer modeling within a shared video diffusion backbone.
    \item Extensive experiments demonstrate that LayerT2V significantly outperforms prior methods in visual fidelity, temporal coherence, and cross-layer consistency. We will release our code and dataset.
\end{itemize}

\section{Related Work}
\label{sec:related_works}

\subsection{Controllable Video Generation}

Video generation has undergone a paradigm shift from adapting 2D architectures to developing native 3D generative frameworks. Early approaches such as Make-A-Video \cite{DBLP:conf/iclr/SingerPH00ZHYAG23}, Tune-A-Video \cite{wu2022tune0a0video0}, as well as AnimateDiff \cite{guo2023animatediff} and VideoCrafter \cite{chen2023videocrafter1}, inflated pretrained T2I backbones (U-Nets/LDMs) with temporal modules (\eg, temporal / cross-frame attention), leveraging static visual priors but often facing temporal artifacts and limited long-horizon coherence.
More recently, Diffusion Transformer \cite{peebles2023scalable} based video generators have emerged, including VDT \cite{DBLP:conf/iclr/LuYFH00D24}, Latte \cite{ma2024latte0}, and trajectory-aware variants such as Tora \cite{zhang2024tora0}. Modern DiT-style systems like CogVideoX \cite{yang2024cogvideox}, Wan \cite{wan2025wan}, and Hunyuan \cite{kong2024hunyuanvideo} treat video latents as spatiotemporal patches with 3D VAEs to model complex motion and details, while Lumiere \cite{bar2024lumiere} introduced a Space-Time U-Net for full-frame-rate generation.
Despite impressive visual quality, these models render foregrounds, backgrounds, and effects into a flattened RGB stream, lacking structural disentanglement and forcing users to regenerate entire scenes for minor edits—unsuitable for professional compositing workflows requiring layer-wise manipulation.

\subsection{Layered Content Generation}
Early studies on layered content generation focused on images. Text2Layer \cite{zhang2023text2layer} and LayerDiffuse \cite{zhang2024transparent} pioneered text-driven RGBA synthesis by modeling transparency, while Alfie \cite{quattrini2024alfie} reduced training costs via inference-time attention optimization. Extending to videos introduces temporal coherence challenges. TransPixeler \cite{wang2025transpixeler}, TransAnimate \cite{chen2025transanimate}, and Wan-Alpha \cite{dong2025video} address this through dedicated alpha tokens, motion modules, and distribution-centric approaches, respectively.
However, these methods generate isolated transparent elements rather than coherent multi-layer videos. Recent efforts such as StM \cite{kara2025stm} and Over++ \cite{qi2025overpp} study layered video composition by explicitly composing foreground and background layers. LayerFlow \cite{ji2025layerflow} is most related, serializing multiple layers along the temporal dimension, but often produces visual disconnection due to data scarcity and lack of explicit inter-layer consistency modeling.

\begin{figure*}[h!]
  \raggedleft
  \includegraphics[width=0.93\textwidth]{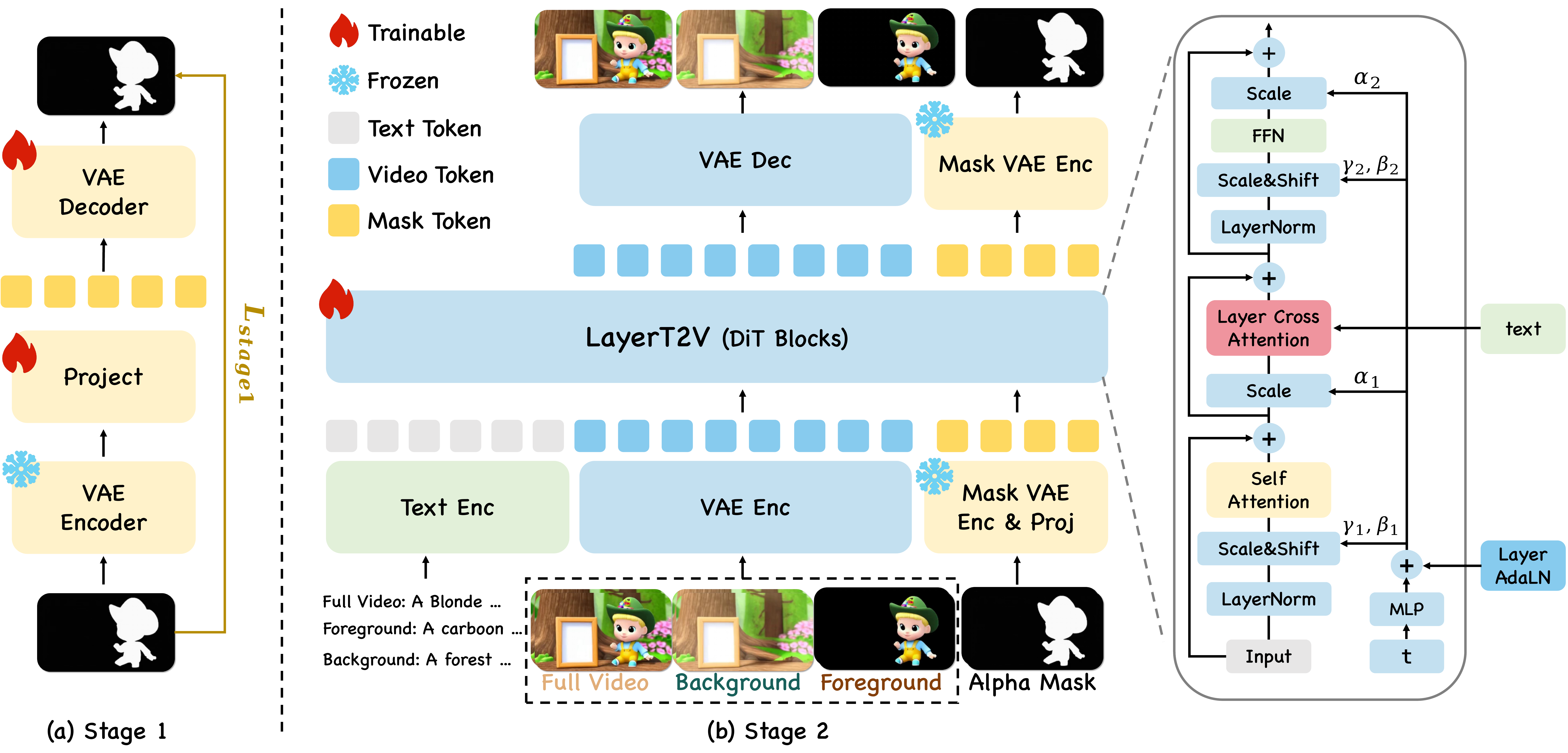}
  \caption{\textbf{Training pipeline and architecture of LayerT2V.} (a) Stage 1: Mask VAE adaptation, where the pretrained VAE encoder is frozen and a lightweight projection plus VAE decoder is trained to reconstruct high-quality alpha mattes. (b) Stage 2: Multi-layer generation with a DiT backbone that jointly models text tokens, video tokens, and mask tokens to generate Full Video, Background, Foreground, and Alpha Mask. LayerAdaLN injects layer identity into the timestep modulation, and layer-aware cross-attention conditions each layer on its corresponding text prompt to improve layer separation and cross-layer coherence.}
  \label{fig:pipe}
  \vspace{-3mm}
\end{figure*}

\section{Method}
\label{sec:method}
This section presents the overall design and training strategy of LayerT2V. We first review the background of Flow Matching in Section~\ref{sec:preliminary}. Then, Section~\ref{sec:pipeline} introduces our unified multi-layer generation pipeline. Finally, Section~\ref{sec:training} details our three-stage training strategy.

\begin{figure*}[t]
  \centering
  \includegraphics[width=0.93\textwidth]{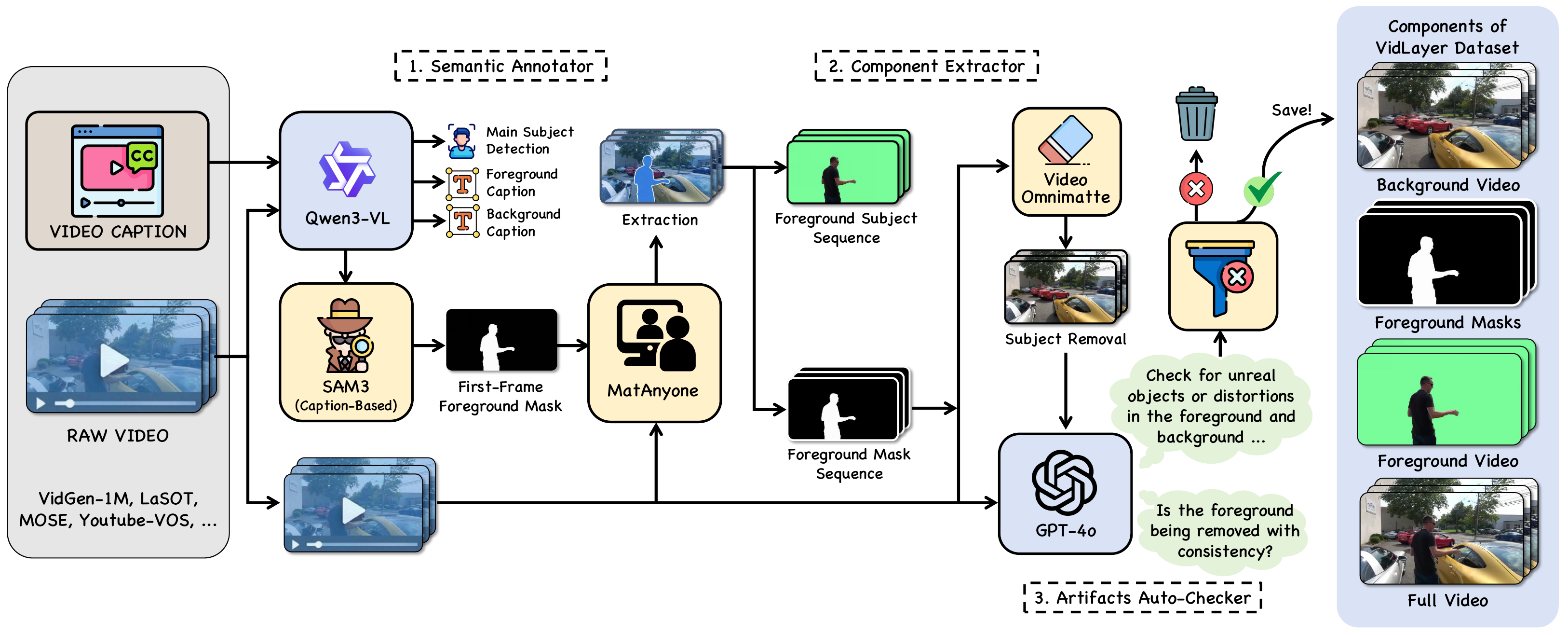}
  \vspace{-2mm}
  \caption{Data construction pipeline of \textit{VidLayer} dataset.}
  \vspace{-3mm}
  \label{fig:data-engine}
\end{figure*}

\subsection{Preliminary}
\label{sec:preliminary}
\paragraph{Flow Matching.}

Flow Matching~\cite{lipman2022flow} learns a time-dependent vector field that transports a prior $p_1$ (\eg, $\mathcal{N}(0, I)$) to the data distribution $p_0$ by minimizing:
\begin{equation} \label{eq:fm_loss_en_final}
\mathcal{L}_{\text{FM}} = \mathbb{E}\!\left[ \,\|v_\theta(x,t) - v_t(x\,|\,x_0,x_1)\|^2 \right],
\end{equation}
where $t\!\sim\!\mathcal{U}(0,1)$, $x_0\!\sim\!p_0$, $x_1\!\sim\!p_1$, $x\!\sim\!p_t(\cdot\,|\,x_0,x_1)$, and $v_t(x\,|\,x_0,x_1)$ is the oracle velocity along a prescribed path.

\paragraph{Rectified Flow.} Rectified Flow~\cite{DBLP:conf/iclr/LiuG023} uses linear paths $x_t = (1-t)x_0 + t x_1$ and learns:
\begin{equation} \label{eq:rf_loss}
\mathcal{L}_{\text{RF}} = \mathbb{E}\!\left[ \,\|v_\theta(x_t,t) - (x_1 - x_0)\|^2 \right],
\end{equation}
which simplifies the oracle velocity and enables efficient ODE integration.

\subsection{LayerT2V Pipeline}
\label{sec:pipeline}

As shown in Figure~\ref{fig:pipe}, LayerT2V jointly models multiple layers within a single generation process. While our framework naturally supports multiple foreground subjects (detailed in Stage 3 of Section~\ref{sec:training}), we present the single-foreground case here for clarity. We first encode each layer video (resolution $H \times W$, $T$ frames) into latent space via a VAE, obtaining latent representations $z_{\text{full}}$, $z_{\text{bg}}$, $z_{\text{fg}}$, and $z_{\text{mask}} \in \mathbb{R}^{C' \times T' \times H' \times W'}$, where $C'$, $T'$, $H'$, and $W'$ denote the channel, temporal, and spatial dimensions in latent space, respectively. We represent the foreground layer as premultiplied content by multiplying the RGB video with the alpha mask before VAE encoding:
\begin{equation}
V_{\text{fg}} = V_{\text{full}} \odot A,\quad z_{\text{fg}} = E(V_{\text{fg}}),
\end{equation}

where $V_{\text{full}}$ is the full video, $A$ is the alpha mask, and $\odot$ denotes
element-wise multiplication with channel broadcasting. We adopt the Wan VAE for RGB layers.
However, alpha masks are single-channel, sparse, and near-binary, with statistics that differ
substantially from natural RGB videos. Directly applying an RGB-pretrained VAE leads to
degraded mask representations and may interfere with joint learning. We therefore fine-tune
the Wan VAE to support masks (see Section~\ref{sec:training}). To construct a unified input
for joint generation, we concatenate the latent codes along the temporal dimension:
\begin{equation}
  z_0 := \mathrm{Concat}\big([z_{\text{full}}, z_{\text{bg}}, z_{\text{fg}},
z_{\text{mask}}]\big),
\end{equation}

where concatenation is performed along the temporal dimension, $z_0 \in \mathbb{R}^{C'\times 4T' \times H' \times W'}$. Temporal concatenation preserves the input structure expected by pretrained video generators, enabling reuse of their temporal modeling capacity without modifying the architecture. Sampling a shared noise $z_1\sim\mathcal{N}(0,I)$ encourages all layers to evolve consistently, improving cross-layer coherence.

Despite these advantages, temporal serialization also introduces a new ambiguity: the backbone must disentangle intrinsic temporal dynamics from the artificial layer ordering. Without explicit layer cues, tokens from different layers may be mixed or misaligned during attention and denoising, harming cross-layer consistency. To address this, we introduce LayerAdaLN and Layered Cross-Attention to explicitly distinguish layer identity.

\paragraph{Layer Adaptive Normalization (LayerAdaLN).}
  Different layers exhibit substantially different distributions: sparse near-binary masks,
  dynamic foregrounds with strong motion, and backgrounds with rich yet largely static
  textures. To introduce layer-specific modulation while sharing backbone parameters, we
  maintain a learnable modulation vector for each layer category $l$:
  \begin{equation}
      \gamma^{(l)} = (b_a^{(l)}, s_a^{(l)}, g_a^{(l)}, b_f^{(l)}, s_f^{(l)}, g_f^{(l)}) \in
  \mathbb{R}^{6 \times d},
  \end{equation}
  which provides shift, scale, and gate parameters for self-attention and FFN. These vectors
  are \textbf{shared across all blocks} and initialized to zero for stable adaptation.

  Given the layer index $l_i$ of token $i$, LayerAdaLN fuses layer modulation with timestep
  modulation $e_t$ via addition: $e_i = e_t + \gamma^{(l_i)}$. Adaptive normalization is then
  applied as:
  \begin{equation}
  \hat{x}=\mathrm{LN}(x)\cdot(1+s)+b,\quad x\leftarrow x+g\cdot\mathcal{F}(\hat{x}),
  \end{equation}
  where $\mathcal{F}$ denotes self-attention or FFN. This design adapts the shared backbone to
  layer-specific statistics with negligible overhead.

\paragraph{Layered Cross-Attention Modulation.}
To achieve fine-grained semantic control while mitigating condition leakage, we modulate the
cross-attention mechanism. We use T5 to independently encode the full-video, foreground, and
background prompts, obtaining context embeddings $c_{\text{full}}$, $c_{\text{fg}}$, and
$c_{\text{bg}}$, which are concatenated along the sequence dimension into a unified context
$c$. This independent encoding prevents semantic leakage across prompts by construction. We construct an attention mask $M \in \mathbb{R}^{L_{\text{vis}}\times L_{\text{text}}}$
to control visibility between visual and text tokens, where $M_{ij}=0$ if text token $j$ is
visible to visual token $i$, and $-\infty$ otherwise. This mask enforces layer-wise semantic
routing: foreground tokens attend only to $c_{\text{fg}}$, background to $c_{\text{bg}}$,
full-video to all three embeddings, and mask tokens follow the foreground pattern. The mask
is injected as an additive bias:
\begin{equation}
\mathrm{Attention}(Q,K,V)=\mathrm{softmax}\!\left(\frac{QK^\top}{\sqrt{d}}+M\right)V,
\end{equation}
where $Q$ is computed from visual tokens, $K$ and $V$ from the text context $c$, and $d$ denotes the attention head dimension.

\begin{figure*}[t!]
  \centering
  \includegraphics[width=0.93\textwidth]{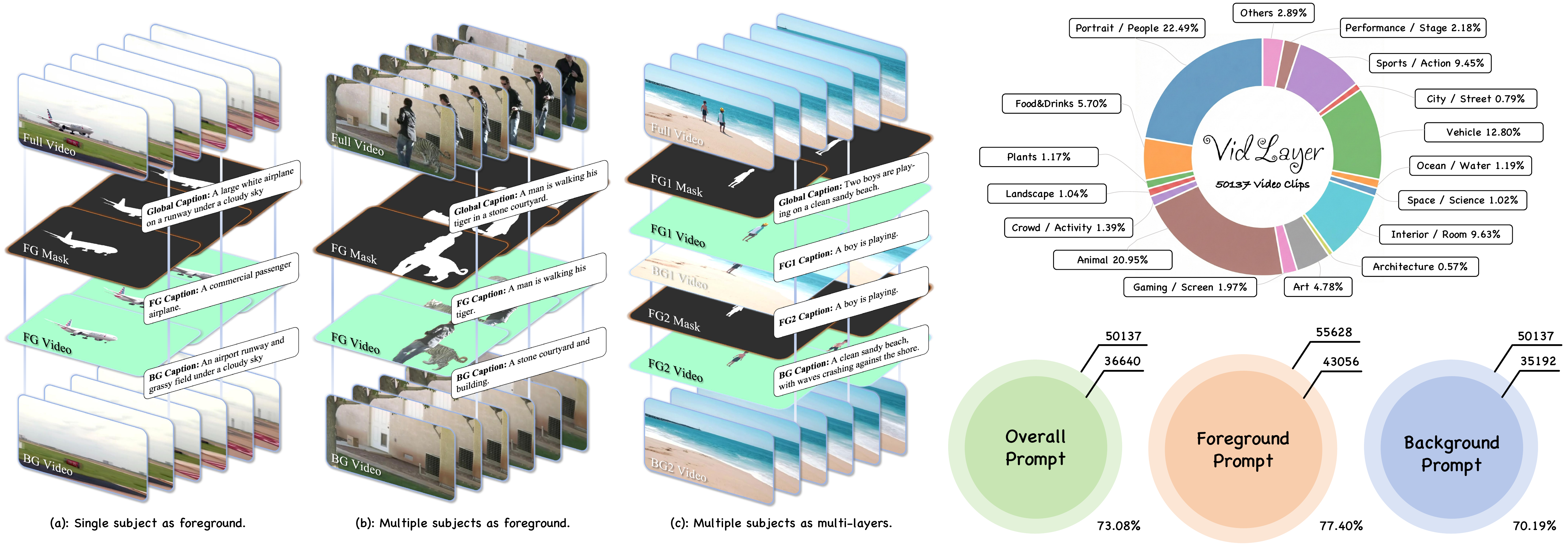}
  \vspace{-2mm}
  \caption{\textbf{Left}: Visualization samples of VidLayer dataset, it involves layered contents and corresponding layered prompts. \textbf{Right}: Scene classifications of VidLayer and semantic redundancy of dataset prompts. For semantic redundancy in text prompts, we extract text embeddings using CLIP~\cite{CLIP} and set a cosine similarity threshold of 0.85 to identify duplicates.}
  \vspace{-3mm}
  \label{fig:data-sample}
\end{figure*}

\subsection{Training and Inference}
\label{sec:training}

LayerT2V adopts a three-stage training strategy to progressively \emph{(i)} adapt the VAE to alpha masks, \emph{(ii)} learn joint multi-layer generation, and \emph{(iii)} multi-foreground extension.

\vspace{-2mm}
\paragraph{Stage 1: Mask VAE Adaptation.}
We adapt the Wan VAE to support mask encoding and decoding by freezing the encoder $E(\cdot)$ and fine-tuning the decoder $D(\cdot)$ with LoRA, while inserting a lightweight projection head after the encoder to adjust the latent mapping. Freezing the encoder preserves the pretrained latent space distribution, ensuring compatibility with the diffusion backbone in Stage 2. Given a ground-truth alpha mask $A\in[0,1]$, we replicate it along the channel dimension to match the RGB input format and optimize:
\begin{equation}
\mathcal{L}_{\text{stage1}} = \mathcal{L}_{\text{rec}} + \lambda_{\text{edge}} \mathcal{L}_{\text{edge}} + \lambda_{\text{perc}} \mathcal{L}_{\text{perc}},
\end{equation}
where $\mathcal{L}_{\text{rec}}$ is a SmoothL1 reconstruction loss, $\mathcal{L}_{\text{edge}}$ enforces boundary sharpness, and $\mathcal{L}_{\text{perc}}$ encourages structural fidelity.

\paragraph{Stage 2: Multi-layer Generation Training.}
We freeze all VAE components and the text encoder, fine-tune the generation backbone, and train the proposed layer-aware modules. Given a multi-layer training sample, we encode each layer and concatenate the latents along the temporal dimension to form the target latent $z_0 \in \mathbb{R}^{C'\times 4T' \times H' \times W'}$. We sample a shared Gaussian noise $z_1\sim\mathcal{N}(0, I)$ with the same shape as $z_0$ and construct the linear path $z_t=(1-t)z_0+t z_1$, where $t\in(0,1)$ is sampled from a logit-normal distribution. The basic objective is the Flow Matching loss:
\begin{equation}
\mathcal{L}_{\text{FM}} = \mathbb{E}_{t,z_0,z_1}\!\left[\left\|v_\theta(z_t,t,c)-(z_1-z_0)\right\|_2^2\right].
\end{equation}

We introduce two auxiliary losses. First, a compositing consistency loss to enforce inter-layer coherence: we recover $\hat{z}_0 = z_t - t\,v_\theta(z_t,t,c)$ and enforce that composed layers match the full video:
\begin{equation}
\mathcal{L}_{\text{cons}} = \left\| \hat{z}_{\text{fg}} + \hat{z}_{\text{bg}} \odot (1-\tilde{A}) - z_{\text{full}} \right\|_2^2,
\end{equation}
where $\tilde{A}$ is the downsampled ground-truth mask. This loss provides a strong supervisory signal for layer consistency. 
Second, to better capture the sparse and near-binary nature of masks, we introduce the mask reconstruction loss:
\begin{equation}
\mathcal{L}_{\text{mask}} = \mathcal{L}_{\text{rec}}(\hat{A}, A) + \lambda_{\nabla}\,\mathcal{L}_{\text{grad}}(\hat{A}, A),
\end{equation}
where $\hat{A}$ is decoded from the predicted mask latent, and $\mathcal{L}_{\text{grad}}$ is a gradient consistency loss that enhances boundary sharpness. The final objective is:
\begin{equation}
\mathcal{L}_{\text{stage2}} = \mathcal{L}_{\text{FM}} + \lambda_{\text{cons}}\,\mathcal{L}_{\text{cons}} + \lambda_{\text{mask}}\,\mathcal{L}_{\text{mask}}.
\end{equation}

\vspace{-2mm}

\paragraph{Stage 3: Multi-foreground Extension.}
We extend LayerT2V to support multiple foreground subjects by serializing additional
foreground-mask pairs along the temporal dimension. The LayerAdaLN module is extended
accordingly, with new foreground layer parameters initialized from the pretrained
single-foreground model to ensure stable adaptation. We continue training on the
multi-subject subset of VidLayer for 5K steps, enabling LayerT2V to generate two
independent foreground layers with their corresponding alpha mattes in a single inference
pass.

\vspace{-2mm}
\paragraph{Inference.}
Given layer-specific text prompts, we first sample a shared Gaussian noise and then perform iterative integration along the ODE trajectory through multiple forward passes. The resulting latent is split along the temporal dimension into individual layer segments, which are decoded by the RGB VAE and Mask VAE respectively to produce the full video, background, foreground, and corresponding alpha matte.
\vspace{-1mm}
\section{VidLayer Dataset}
\label{sec:vidlayer}

\begin{figure*}[t]
  \centering
  \includegraphics[width=0.9\textwidth]{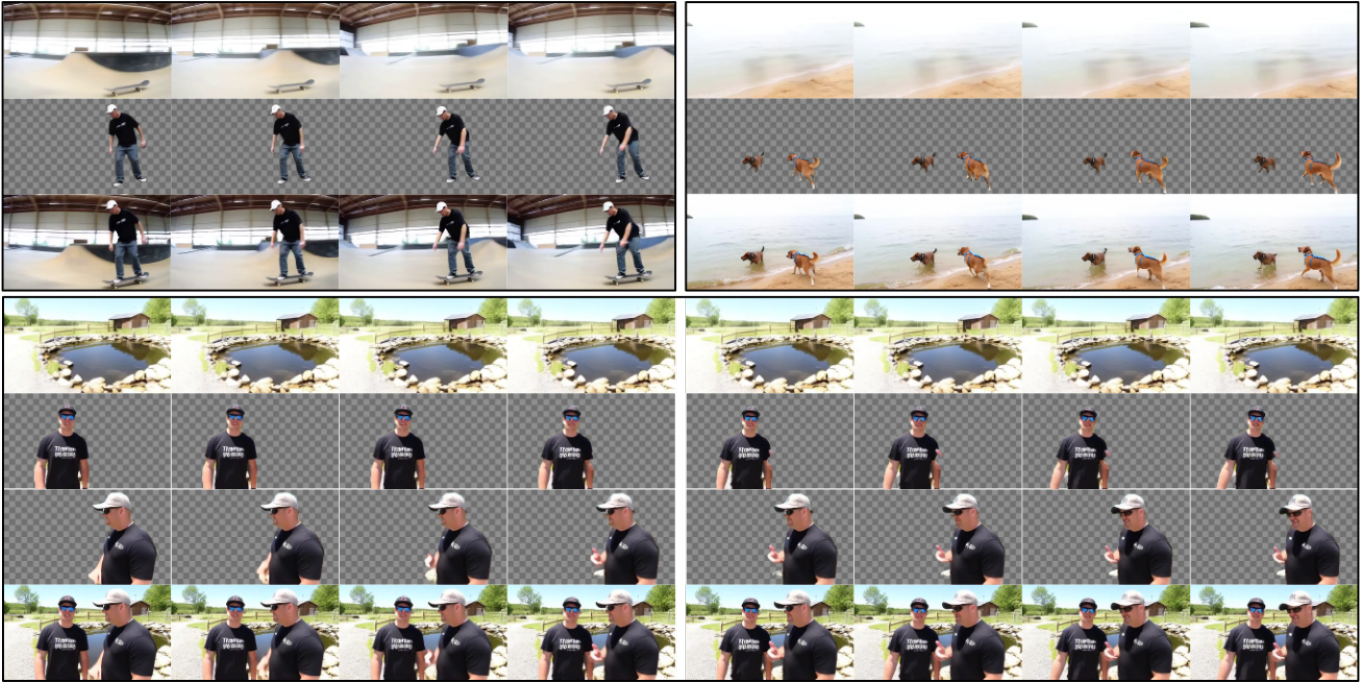}
  \caption{\textbf{Qualitative results.} LayerT2V generates high-fidelity multi-layer videos across three generation modes: (a) single-foreground with a single subject, (b) single-foreground with multiple subjects, and (c) multi-foreground joint generation with independent layers. Our method produces clean foreground separation, sharp alpha mattes, and complete backgrounds without leakage or boundary artifacts across diverse scenes and motion patterns.}
  \label{fig:qual}
  \vspace{-3mm}
\end{figure*}

\vspace{-1mm}
Recent progress in T2V generation has largely focused on synthesizing visually coherent full-frame videos. However, most existing benchmarks and training datasets only provide monolithic video representations, without explicitly modeling the internal compositional structure of a scene. 

To support our proposed LayerT2V model, we introduce \textbf{VidLayer}, a large-scale dataset specifically designed for layer-aware video generation. VidLayer provides aligned foreground videos, foreground masks, background videos, and raw videos, enabling supervision at both the semantic and structural levels. Since such data is not available in the community, constructing VidLayer is a foundational step toward layer-wise text-to-video modeling.

\vspace{-2mm}
\subsection{Data Construction Pipeline}
As illustrated in Figure~\ref{fig:data-engine}, VidLayer is constructed through an automated data engine pipeline consisting of three stages: \emph{(i) semantic annotation}, \emph{(ii) multi-layer component extraction}, and \emph{(iii) automatic quality review}. We start from raw caption--video pairs sampled from multiple community datasets and progressively transform them into structured multi-layer video representations. The entire pipeline is fully automated and scalable, allowing us to curate high-quality data at scale without manual intervention.

\vspace{-2mm}
\subsection{Semantic Annotation}
We first collect over 200k videos from the VidGen~\cite{tan2024vidgen}, LaSOT~\cite{fan2019lasot}, MOSE~\cite{ding2023mose}, Youtube-VOS~\cite{xu2018youtube}, GOT-10k~\cite{huang2019got} datasets, each paired with a natural language caption. Using the video understanding capability of Qwen3-VL~\cite{Qwen3-VL}, we perform \textit{subject-oriented semantic annotation} for each video. Specifically, the annotator identifies the primary foreground subject (\eg, a person or a salient object) and generates two complementary textual descriptions: a \textit{foreground caption} describing the subject itself, and a \textit{background caption} characterizing the remaining scene context.

\vspace{-1mm}
To extract foreground subjects, we further exploit the caption-to-mask capability of SAM3~\cite{carion2025sam}. By combining the foreground caption with the first video frame, SAM3 produces an initial foreground mask. This mask serves as a reliable spatial prior for subsequent temporally consistent video segmentation.

\vspace{-2mm}
\subsection{Multi-layer Component Extraction}
Based on the first-frame foreground mask, we employ MatAnyone~\cite{yang2025matanyone} to extract temporally consistent multi-layer components. Given the original video and the initial mask, MatAnyone outputs a sequence of \textit{foreground masks} as well as a corresponding \emph{foreground subject video}, where the subject is rendered on a green-screen background. This process robustly handles challenging scenarios such as occlusion, pose variation, and scale changes, ensuring temporal coherence of the extracted subject. We then apply Gen-Omnimatte~\cite{lee2025generative}, a video matting method for \emph{foreground removal}. 
The resulting background preserves realistic spatial structures and temporal continuity without introducing noticeable artifacts.

At the end of this stage, each video is decomposed into a structured set of aligned components: a foreground video, a foreground mask sequence, a background video, and the original full video. We provide additional examples in Figure~\ref{appendix:dataset_sample} for enhanced illustration.

\subsection{Automatic Quality Control}
Although the pipeline is fully automated, naive application of video decomposition models can introduce various failure cases. To ensure the reliability and usability of VidLayer, we employ GPT-4o~\cite{openai2024gpt4technicalreport} as an \emph{Artifacts Auto-Checker} to perform strict quality control on the generated components. 
GPT-4o evaluates multiple criteria, including whether the foreground subject is clear and recognizable, and whether the background videos involve the presence of unrealistic objects, structural distortions, or incomplete human limbs caused by inpainting artifacts. Only samples that pass all quality checks are retained in the final dataset.

\subsection{Dataset Statistics and Properties}
After automatic filtering, the resulting \textbf{VidLayer} dataset contains \textbf{50K} high-quality video clips, comprising approximately \textbf{4M} frames in total. Among them, around 0.6M frames correspond to multi-subject scenarios involving two or three foreground subjects. Each sample in VidLayer provides aligned multi-layer representations, enabling direct supervision for layer-aware video generation, decomposition, and editing tasks. We also calculate \textit{semantic redundancy} in text prompts and scene classifications to ensure diversity, as shown in Figure~\ref{fig:data-sample}.

VidLayer fills a critical gap in existing video datasets by offering large-scale, structured multi-layer video data, and serves as a key enabler for LayerT2V model and future research on compositional and controllable video generation.

\begin{figure*}[t]
  \centering
  \includegraphics[width=0.9\textwidth]{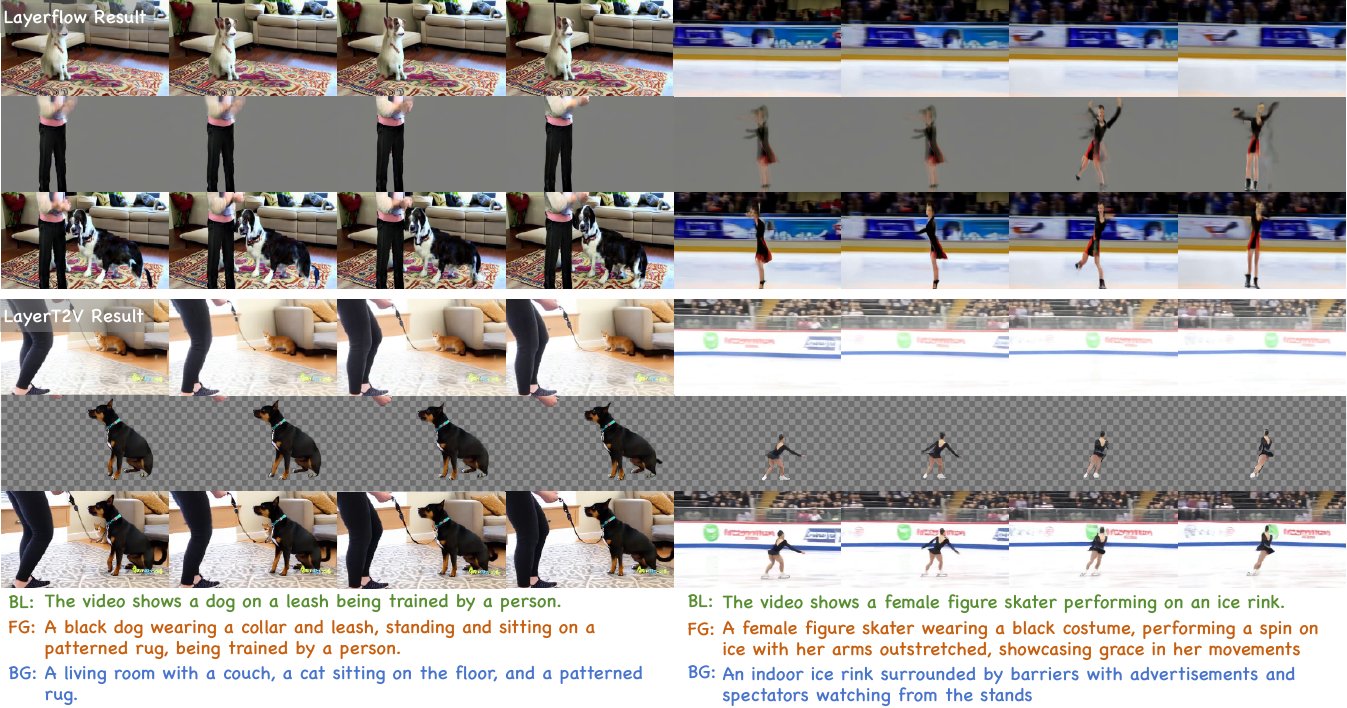}
  \caption{\textbf{Qualitative comparison.} Compared to LayerFlow, LayerT2V produces higher-quality video layers with stronger temporal consistency and better text alignment. BL/BG/FG correspond to the full-video/background/foreground prompts. Note that LayerFlow outputs the foreground as RGB (without alpha), as it claims RGB foregrounds achieve higher visual quality than RGBA.}
  \label{fig:qual_comp}
  \vspace{-3mm}
\end{figure*}

\section{Experiments}
\label{sec:exp}

\subsection{Experimental Setup}
\label{sec:exp_setup}

  \begin{table*}[h!]
      \centering
      \caption{\textbf{Quantitative comparison on VBench.} We evaluate five dimensions across
  foreground (FG), background (BG), and full-video (BL) layers on 200 held-out prompts from
  VidLayer. Higher values indicate better performance. Best results are highlighted in \textbf{bold}.}
      \label{tab:vbench}
      \renewcommand{\arraystretch}{1.3}  
      \resizebox{\linewidth}{!}{
      \begin{tabular}{l|ccc|ccc|ccc|ccc|ccc}
      \toprule
      \multirow{2}{*}{Method} & \multicolumn{3}{c|}{Aesthetic Quality $\uparrow$} &
      \multicolumn{3}{c|}{Motion Smoothness $\uparrow$} & \multicolumn{3}{c|}{Temporal Flickering $\uparrow$} &
      \multicolumn{3}{c|}{Subject Consistency $\uparrow$} & \multicolumn{3}{c}{Text Alignment $\uparrow$} \\
      \cmidrule(lr){2-4} \cmidrule(lr){5-7} \cmidrule(lr){8-10} \cmidrule(lr){11-13}
      \cmidrule(lr){14-16}
      & FG & BG & BL & FG & BG & BL & FG & BG & BL & FG & BG & BL & FG & BG & BL \\
      \midrule
      LayerFlow & 0.4591 & 0.4534 & 0.4895 & 0.9582 & 0.9891 & 0.9788 & 0.9630 & 0.9736 &
  0.9634 & 0.9440 & 0.9682 & 0.9624 & 0.1727 & 0.1849 & 0.1941 \\
      LayerT2V (Native Mask VAE) & 0.4432 & 0.5327 & 0.5010 & 0.9863 & 0.9908 & 0.9776 & 0.9661
   & \textbf{0.9837} & 0.9669 & 0.9828 & \textbf{0.9839} & 0.9749 & 0.1846 &
  0.2102 & 0.1996 \\
     \rowcolor{blue!9} \textbf{LayerT2V (VAE LoRA)} & \textbf{0.4971} & \textbf{0.5577} &
  \textbf{0.5391} & \textbf{0.9919} & \textbf{0.9920} &
  \textbf{0.9845} & \textbf{0.9872} & 0.9830 & \textbf{0.9676} &
  \textbf{0.9829} & 0.9838 & \textbf{0.9750} & \textbf{0.2009} &
  \textbf{0.2307} & \textbf{0.2136} \\
      \bottomrule
      \end{tabular}
      }
\renewcommand{\arraystretch}{1.0}  
  \vspace{-10pt}
  \end{table*}

\paragraph{Implementation Details.}
LayerT2V is built upon the pretrained Wan2.1-1.3B~\cite{wan2025wan} video generation model. We use the AdamW optimizer with a learning rate of $1 \times 10^{-4}$ and a cosine schedule with 500 warmup steps. All experiments are conducted on 8 NVIDIA H20 GPUs. In Stage 1, we freeze the VAE encoder and fine-tune the decoder with LoRA (rank 128). In Stage 2, we insert LoRA adapters along with the proposed layer-aware modules into the pretrained backbone. In Stage 3, we continue training on the multi-subject subset of VidLayer to support up to three foreground layers. All stages adopt a progressive training strategy from low to high resolution. Detailed hyperparameters are provided in the appendix~\ref{sup:implement}.
\vspace{-2mm}
\paragraph{Evaluation Metrics.}
We adopt VBench~\cite{huang2023vbench0} as our primary evaluation framework, which provides a comprehensive assessment across multiple dimensions, including subject consistency, temporal flickering, motion smoothness, and aesthetic quality. For text alignment, we use ViCLIP~\cite{wang2023internvid0} to measure the semantic correspondence between generated videos and input prompts. We compare LayerT2V against \textbf{LayerFlow}~\cite{ji2025layerflow}, a recent multi-layer video generation approach. To ensure fair evaluation, we carefully sample 200 layered prompts from the VidLayer dataset and remove them from our training set. All quantitative results are reported on this fixed held-out set.
\subsection{Analysis and Discussion}
\label{sec:analysis}
\paragraph{Qualitative Results.}

Fig.~\ref{fig:qual} visualizes LayerT2V across three generation modes: single-foreground with a single subject, single-foreground with multiple subjects, and multi-foreground joint
generation with independent layers. Even under fast motion and complex styles, our method
produces clean foregrounds with sharp, temporally stable alpha mattes and complete
backgrounds without foreground leakage or boundary flickering. Fig.~\ref{fig:qual_comp}
presents the qualitative comparison results: LayerT2V shows consistently higher-quality
layers and more consistent recompositions.
\vspace{-2mm}
\paragraph{Quantitative Evaluation.} 
We adopt VBench~\cite{huang2023vbench0} to evaluate foreground (FG), background (BG), and blending results (BL) across five dimensions: \textit{Aesthetic Quality}, \textit{Motion Smoothness}, \textit{Temporal Flickering}, \textit{Subject Consistency}, and \textit{Text Alignment}. As shown in Table~\ref{tab:vbench}, LayerT2V achieves strong performance across all metrics. The FG exhibits high subject consistency with minimal boundary jitter, while the BG remains temporally coherent without foreground leakage. The recomposited BL preserves favorable scores, validating both individual layer quality and cross-layer coherence.

\begin{table}[t]
      \centering
      \caption{\textbf{User study results.} Values indicate the fraction of times each
      method is selected as the best (higher is better). Best results are highlighted in \textbf{bold}.}
      \label{tab:userstudy}
      \vspace{-2pt}
      \resizebox{\linewidth}{!}{
      \begin{tabular}{l|ccc}
        \toprule
        Method & Aesthetic $\uparrow$ & FG Quality $\uparrow$ & Text Align. $\uparrow$ \\
        \midrule
        LayerFlow & 0.120 & 0.156 & 0.136  \\
        LayerT2V (Native Mask VAE) & 0.156 & 0.076 & 0.196  \\
        \rowcolor{blue!9} \textbf{LayerT2V (Ours)} & \textbf{0.724} & \textbf{0.768} &
  \textbf{0.668}  \\
        \bottomrule
      \end{tabular}
      }
      \vspace{-4pt}
  \end{table}

  \begin{table}[t]
    \centering
    \caption{\textbf{Ablation study on layer-aware modules.} Baseline: vanilla temporal concatenation
  only. \textcircled{1}: LayerAdaLN,
    \textcircled{2}: LayerCrossAttention. Best results are highlighted in \textbf{bold}.}
    \label{tab:ablation_modules}
    \vspace{-2pt}
    \renewcommand{\arraystretch}{1.3}  
    \resizebox{\linewidth}{!}{
    \begin{tabular}{l|ccc|ccc|ccc}
    \toprule
    \multirow{2}{*}{Configuration} & \multicolumn{3}{c|}{Subject Consistency$\uparrow$}
    & \multicolumn{3}{c|}{Temporal Flickering$\uparrow$} & \multicolumn{3}{c}{Text
    Alignment$\uparrow$} \\
    \cmidrule(lr){2-4} \cmidrule(lr){5-7} \cmidrule(lr){8-10}
    & FG & BG & BL & FG & BG & BL & FG & BG & BL \\
    \midrule
    Baseline & 0.931 & 0.942 & 0.924 & 0.955 & 0.961 & 0.945 & 0.169 & 0.182 & 0.188 \\
    Baseline + \textcircled{1} & 0.973 & 0.975 & 0.961 & 0.979 & 0.976 & 0.959 & 0.180 & 0.194
  & 0.195 \\
    \rowcolor{blue!9} \textbf{Baseline + \textcircled{1} + \textcircled{2}} & \textbf{0.983} &
  \textbf{0.984} & \textbf{0.975} & \textbf{0.987} &
  \textbf{0.983} & \textbf{0.968} & \textbf{0.201} &
  \textbf{0.231} & \textbf{0.214} \\
    \bottomrule
    \end{tabular}
    }
    \renewcommand{\arraystretch}{1.0}  
    \vspace{-10pt}
  \end{table}
\vspace{-2mm}
\paragraph{User Study.}

We aslo conduct a user study to evaluate the perceptual quality of generated multi-layer videos. Participants are asked to compare results from LayerFlow, LayerT2V with Native Mask VAE, and our LayerT2V across three aspects: Aesthetic Quality, Foreground Quality and Text Alignment. As shown in Table~\ref{tab:userstudy}, LayerT2V is consistently preferred over all baselines, demonstrating superior layer separation and visual quality.

\subsection{Ablation Study.}
\label{sec:ablation}

\begin{figure}[t!]
  \centering
  \includegraphics[width=0.9\linewidth]{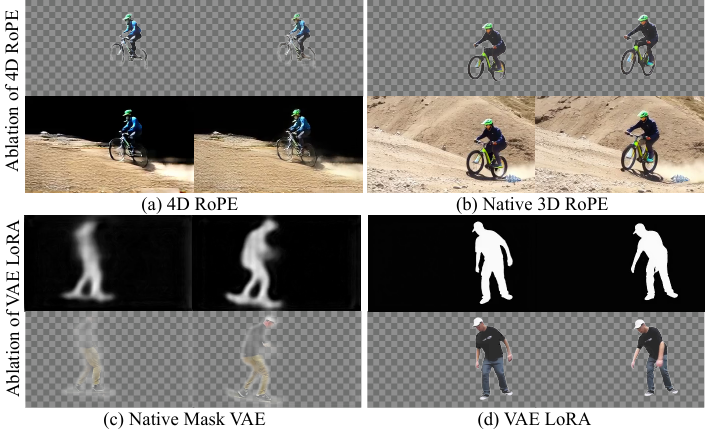}
  \caption{\textbf{Ablation results.} (a-b) Using 4D RoPE disrupts the pretrained spatiotemporal positional encoding, severely degrading generation quality. (c-d) Compared to the Native Mask VAE, our VAE LoRA strategy produces higher-quality foreground masks.}
  \label{fig:ablation}
  \vspace{-4mm}
\end{figure}

\paragraph{4D RoPE for Layer-aware Position Embedding.}
A natural idea is to extend the standard 3D RoPE~\cite{su2021roformer0}, which operates over the temporal and spatial dimensions, to a 4D variant by introducing an additional layer axis. However, as shown in Fig.~\ref{fig:ablation} (a) and (b), this proves ineffective: rather than promoting disentanglement, 4D RoPE interferes with the pretrained positional embedding, causing inter-frame flickering and boundary artifacts. We attribute this to the fact that layer identity is a categorical attribute rather than a spatial or temporal dimension—positional embedding alone cannot disentangle the artificial layer ordering from natural temporal dynamics, necessitating dedicated layer conditioning mechanisms.

\paragraph{Alpha Mask Processing.} We compare two strategies: training a native Mask VAE from scratch versus our VAE LoRA method. As shown in Fig.~\ref{fig:ablation} (c) and (d), the native Mask VAE produces discernible layer decomposition, validating that treating alpha masks distinctly from RGB is feasible. However, it exhibits significant quality deficiencies: temporal flickering, blurry boundaries, and inconsistent alpha values, resulting in poor mask fidelity. We attribute these artifacts to the lack of pretrained video priors. In contrast, VAE LoRA leverages the spatial and temporal representations of the pretrained Wan VAE, applying minimal trainable parameters to extend the decoder for mask reconstruction while preserving RGB encoding quality.

\paragraph{Layer-Aware Modules.} Table~\ref{tab:ablation_modules} validates that both LayerAdaLN and Layered Cross-Attention are essential. The baseline (temporal concatenation only) suffers from foreground-background entanglement, low subject consistency and poor text alignment due to semantic leakage. Adding LayerAdaLN improves layer separation and temporal stability, yet text alignment remains suboptimal as cross-attention still allows inter-layer contamination. The full model achieves the best results: LayerAdaLN ensures clean decomposition, while Layered Cross-Attention enforces strict visibility constraints, maximizing consistency and alignment across all layers.

\section{Conclusion}
\label{sec:conclusion}

We present \textbf{LayerT2V}, a unified multi-layer video generation framework that produces semantically consistent full videos, background layer, and multiple foreground RGB layers with corresponding alpha mattes in a single inference pass. To address layer-identity ambiguity and conditional leakage, we design LayerAdaLN and layer-aware cross attention modulation. We also release \textbf{VidLayer}, the first large-scale multi-layer video dataset. Experiments demonstrate that LayerT2V substantially outperforms prior methods in visual fidelity, temporal consistency, and cross-layer coherence. We hope this work advances multi-layer video generation and enables fine-grained control in professional video production.

\bibliography{main}
\bibliographystyle{main}

\newpage
\appendix
\onecolumn

\section{Implementation Details}
\label{sup:implement}
LayerT2V is built upon the pretrained Wan2.1-1.3B~\cite{wan2025wan} video generation
model. We use the AdamW optimizer with a learning rate of $1 \times 10^{-4}$, weight
decay of 0.01, and a cosine learning rate schedule with 500 warmup steps. Timesteps
are sampled from a logit-normal distribution with mean 0.5 and standard deviation
1.0. All experiments are conducted on 8 NVIDIA H20 GPUs.

\textit{Stage 1: Mask VAE Adaptation.}
We freeze the Wan VAE encoder and fine-tune the decoder with LoRA (rank 128). The loss weights are set to $\lambda_{\text{edge}} = 1.0$ and $\lambda_{\text{perc}} = 1.0$. We adopt a progressive training strategy: first training at $192 \times 336$ resolution with 9 frames (batch size 8, 4.5K steps), then scaling to $384 \times 672$ with 41 frames (batch size 2, 2.5K steps).

\textit{Stage 2: Multi-layer Generation.}
Starting from the pretrained Wan2.1 backbone, we insert LoRA adapters (rank 128) along with the proposed LayerAdaLN and layer-aware cross-attention modules. Loss weights are set to $\lambda_{\text{cons}} = 0.1$, $\lambda_{\text{mask}} = 0.1$, and $\lambda_{\nabla} = 0.1$. We first train at $192 \times 336$ with 9 frames (batch size 8, 6K steps), then fine-tune at $384 \times 672$ with 41 frames (batch size 4, 1K steps).

\textit{Stage 3: Multi-foreground Extension.}
Using the same training configuration as the high-resolution phase of Stage 2 ($384 \times 672$, 41 frames, batch size 1), we continue training for 5K steps. This stage extends the model to support up to three independent foreground layers with their corresponding alpha mattes.

\section{Computational Cost}
\label{sup:cost}

We analyze the inference efficiency of LayerT2V on a single NVIDIA H200 GPU (141GB memory).
Table~\ref{tab:cost} summarizes the computational cost for generating multi-layer videos at
$384 \times 672$ resolution.

\begin{table}[h!]
  \centering
  \caption{\textbf{Inference cost of LayerT2V.} Measured on a single NVIDIA H200 GPU for
41-frame video generation with single-foreground configuration.}
  \label{tab:cost}
  \vspace{2pt}
  \begin{tabular}{l|c}
      \toprule
      \textbf{Metric} & \textbf{Value} \\
      \midrule
      Resolution & $384 \times 672$ \\
      Output frames per layer & 41 \\
      Total output frames & 164 (Full + BG + FG + Mask) \\
      Inference time & 2 min 31 sec \\
      Peak GPU memory & 24 GB \\
      \bottomrule
  \end{tabular}
\end{table}

\section{Data Construction Details}

\subsection{Semantic Annotation and Components Extraction}

We use Qwen3-VL-30B-A3B-Instruct~\cite{Qwen3-VL} to detect main foreground subject and obtain layer-aware prompts. The prompt we use for this stage is visualized in the following prompt box:

\begin{promptBox}{System Message for Qwen3-VL-30B-A3B-Instruct} 
\begin{lstlisting}
You are a video content analysis assistant.

Analyze the video and output STRICT JSON ONLY.

Tasks:
1. Find 1 to 2 main subjects
2. Determine if the main subject is always visible
3. Estimate subject size category
4. Determine camera stability
5. Rate overall video quality

Rules for subjects:
- Subject must be a single object (not a group)
- Subject occupies at least 10% of the frame
- Subject occupies at most 70% of the frame
- If no valid subject exists, output empty subjects list

Definitions:
- subject_size must be one of: "small", "medium", "large"
- video_quality is an integer from 0 to 10

Output format:
{
  "subjects": [
    {
      "subject_id": "<string>",
      "subject_description": "<string>"
    }
  ],
  "subject_visibility": {
    "always_visible": true,
    "subject_size": "medium"
  },
  "camera_stability": {
    "stable": true
  },
  "video_quality": 7
}
\end{lstlisting}
\end{promptBox}

\begin{figure}[h!]
    \centering
    \includegraphics[width=0.98\linewidth]{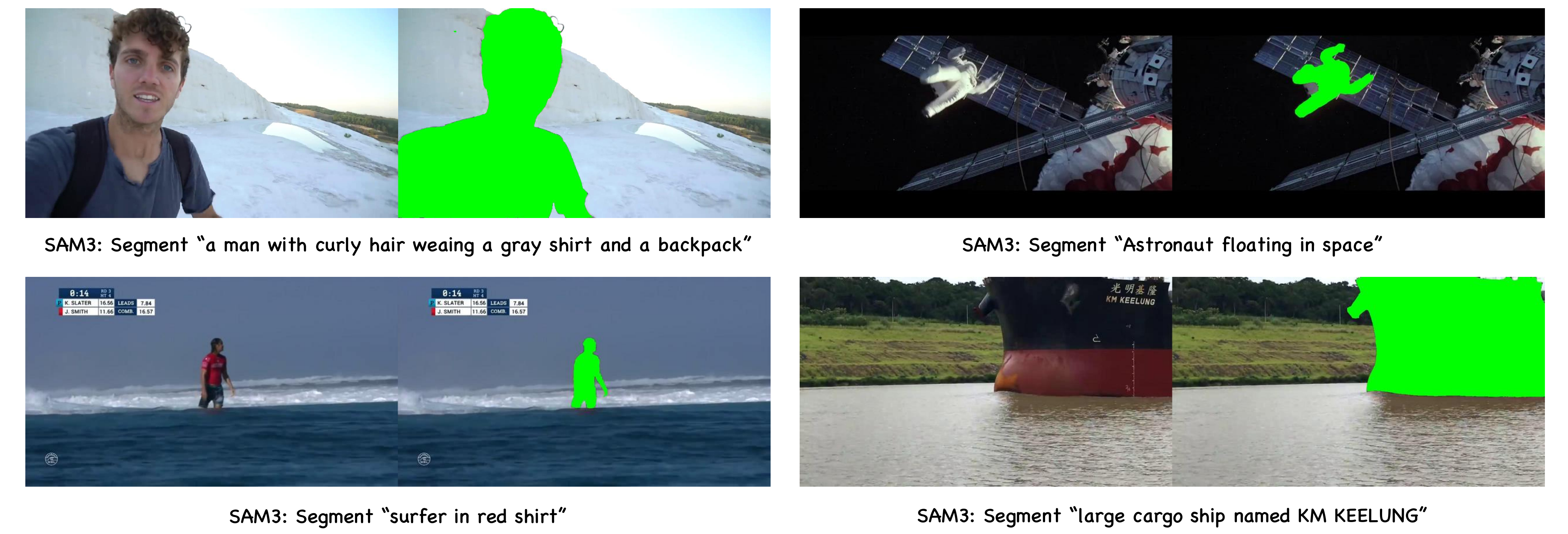}
    \caption{SAM3~\cite{carion2025sam} takes foreground caption as input, and output its local region with binary mask.}
    \label{appendix:sam3}
\end{figure}

Then prompt-to-mask capability of SAM3 is utilized to extract first-frame mask of the video foreground, as visualized in Figure~\ref{appendix:sam3}, as a reliable cornerstone for the following sequential masks extraction by MatAnyone~\cite{yang2025matanyone}.

\subsection{Details for Artifacts Auto Check}

\begin{figure}[h!]
    \centering
    \includegraphics[width=0.98\linewidth]{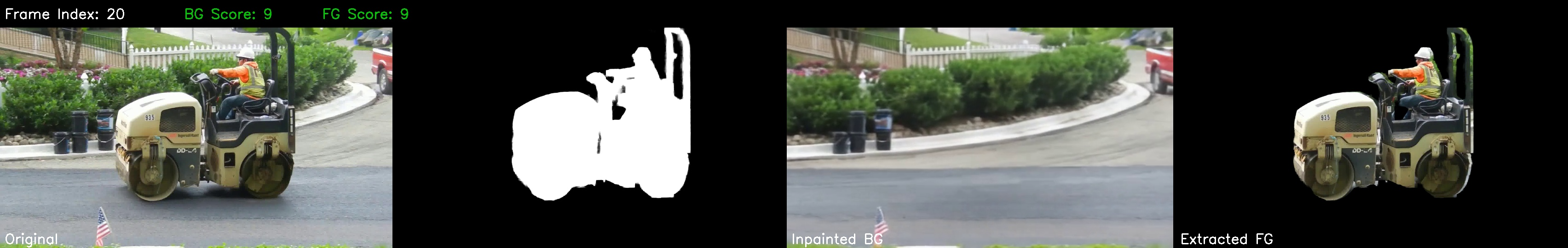}
    \includegraphics[width=0.98\linewidth]{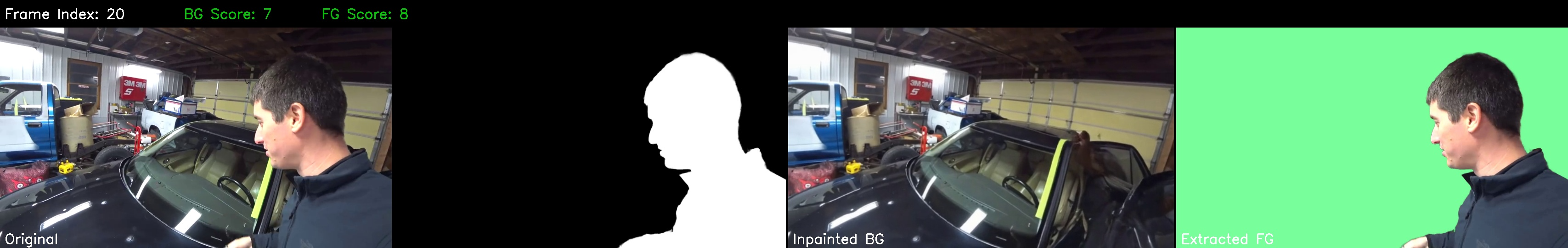}
    \includegraphics[width=0.98\linewidth]{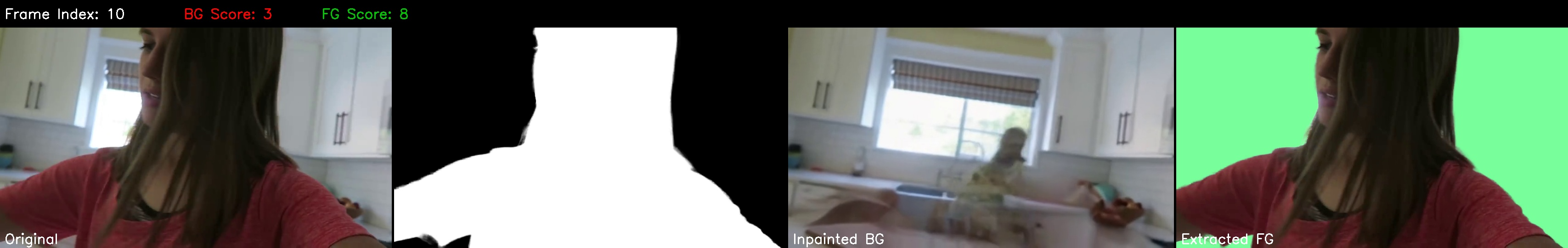}
    \caption{GPT-4o quality check results. In the first two rows, the foreground is removed with excellent consistency. But in the third row, there appears some texture leakage from the foreground to the background. This phenomenon can be attributed to the drawbacks of the Gen-Omnimatte~\cite{lee2025generative} model. GPT-4o will label such cases as low scores and remove them from the dataset.}
    \label{appendix:gpt4o}
\end{figure}

To ensure the visual fidelity and decomposition quality of the VidLayer dataset, we implement an automated quality assurance pipeline leveraging the multimodal capabilities of GPT-4o~\cite{openai2024gpt4technicalreport}. For each sampled key-frame, we construct \textbf{a composite visual prompt} by horizontally concatenating the original video frame, the object mask, the inpainted background, and the extracted foreground, enabling the model to perform direct reference-based comparisons, as demonstrated in Figure~\ref{appendix:gpt4o}. We employ a rigorous scoring rubric that strictly penalizes hallucinations—such as “ghosting” artifacts, color leakage, and geometric distortions—while distinguishing plausible scene interactions (\textit{e.g.}, cast shadows) from unwanted residues, quantifying quality via dual metrics for background reconstruction and foreground extraction. 

Ultimately, temporal segments are validated based on the consistency of high scores across adjacent key-frames, where only sequences demonstrating superior layer separation are cropped and incorporated into the final dataset. For example, we take intervals between key-frames as $\Delta_{key} = 10$, then if both \textit{BG Score} and \textit{FG Score} of frame $N$ and frame $N+\Delta_{key}$ are $\geq 7$, we consider frame interval $[N,N+\Delta_{key}]$ as qualified. The system prompt we use for this stage is given in the following prompt box:

\begin{promptBox}{System Message for GPT-4o} 
\begin{lstlisting}
You are a Data Quality Expert for a Layered Video Generation Model.
Your goal is to curate a training dataset where a video is decomposed into:
1. [BG] Background Layer.
2. [FG] Foreground Layer.

Context Description: {description}

### INPUT LAYOUT
Each image consists of **4 horizontally concatenated panels** (Left to Right):
- **Panel 1 [Original]**: The ground truth video frame.
- **Panel 2 [Mask]**: White area indicates the foreground object location.
- **Panel 3 [BG]**: The inpainted background result (The object should be gone).
- **Panel 4 [FG]**: The extracted foreground result (The object on black).

### VISUAL COMPARISON STRATEGY
**Step 1: Evaluate Foreground (fg_score)**
- Compare Panel 4 [FG] directly against Panel 1 [Original].
- Check: Sharpness, edge accuracy, and semantic preservation.

**Step 2: Evaluate Background (bg_score) - CRITICAL STEP**
- Look at Panel 2 [Mask] to find the region of interest.
- **COLOR LEAKAGE CHECK**: Look at the color of the person in Panel 1 (e.g., Red shirt, Brown hair). Now look at Panel 3 [BG] in that same spot. 
    - Is there a blurry blob of the **SAME COLOR** (e.g., a red or brown smudge) floating there? 
    - If YES -> This is "Color Leakage/Ghosting", NOT a shadow. Score must be **0-3**.
- **SHADOW VS. BLOB CHECK**: 
    - Real shadows are usually dark/grey and cast on the floor/wall. 
    - **Floating, colored, amorphous clouds** are artifacts.

### SCORING RUBRIC
**BG SCORING (First Value)**
- **0-3 (FATAL - REJECT)**: 
    * **Ghosting/Color Leakage**: A blurry color blob that matches the foreground object's color.
    * **Floating Blobs**: Large amorphous shapes floating in the middle of the room.
    * **Paridolia**: Human features (eyes, face) appearing in the foreground mask region (outside mask region is acceptable).
    * **Melting textures**.
- **4-6 (Mediocre)**: Blurry, smudged, visible clone-stamping, but no scary ghosts.
- **7-9 (Good)**: Clean. 
    * **ACCEPTABLE**: Sharp/Real props (strings, balls).
    * **ACCEPTABLE**: **Cast Shadows** (Must be dark/grey and geometrically consistent on surfaces).
- **10 (Perfect)**: Flawless.

**FG SCORING (Second Value)**
- **0-3 (Poor)**: Unrecognizable blob, heavy noise.
- **4-6 (Fair)**: Blurry, rough edges.
- **7-9 (Good)**: Clear, recognized, slight halos.
- **10 (Perfect)**: Crystal clear.

### OUTPUT FORMAT
Return a strictly valid JSON object. 
Keys: Frame Index (string).
Values: **A list of exactly two integers: [bg_score, fg_score]**.

Example: {{ "0": [9, 8], "10": [2, 9], "20": [10, 10] }}
\end{lstlisting}
\end{promptBox}

\newpage
\begin{figure}[t!]
    \centering
    \includegraphics[width=0.95\linewidth]{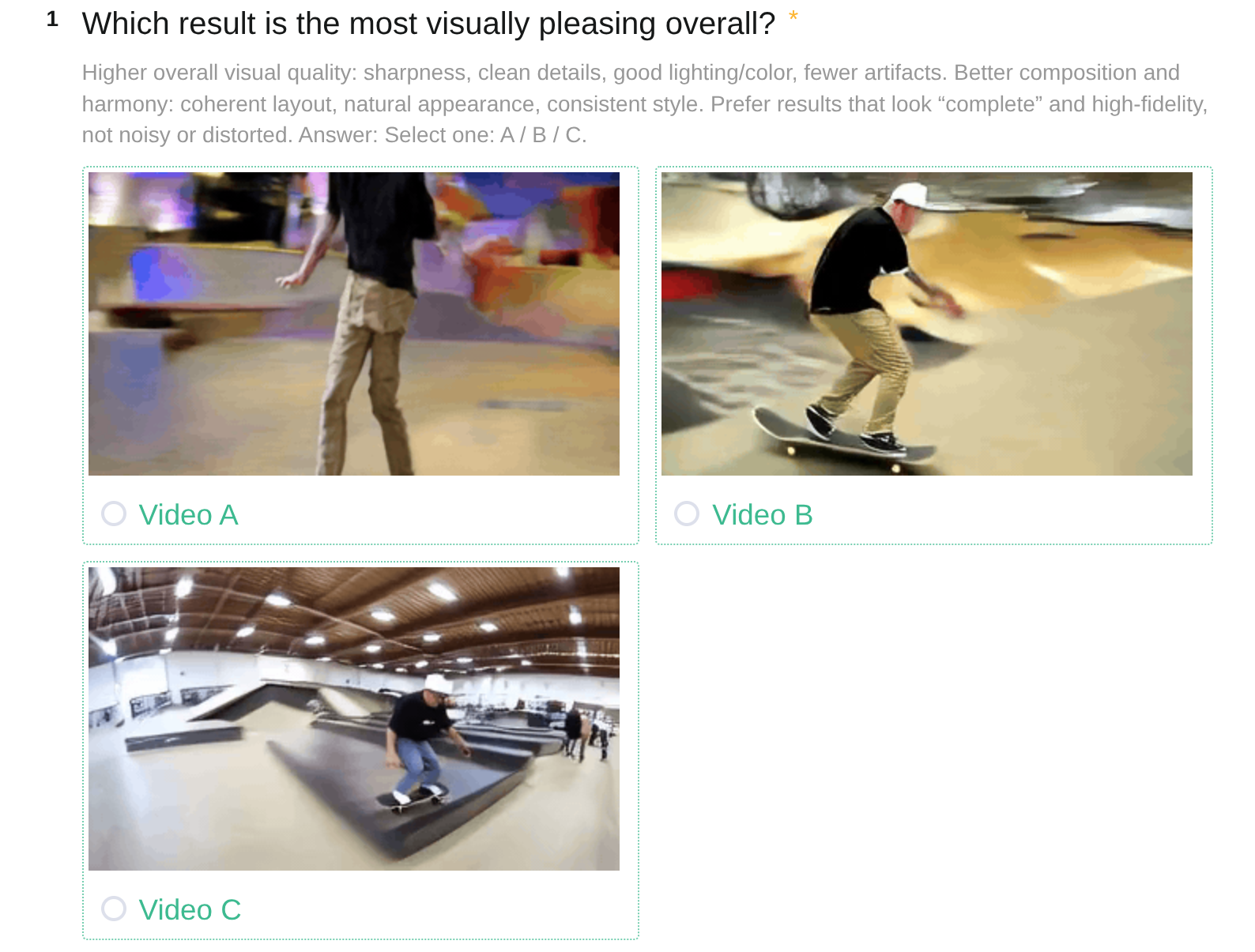}
    \caption{\textbf{User study interface.} Participants are shown video results from three methods in randomized order and asked to select the best one for each evaluation aspect.}
    \label{fig:userstudy}
\end{figure}
\section{User Study}

  \paragraph{Study Design.} We conduct a user study to evaluate perceptual quality aspects that
   automated metrics may not fully capture. We recruit 30 participants, including both individuals with video editing experience and general users. We randomly sample 25 prompts from the VidLayer test set, covering diverse subjects including humans, animals, and objects. For each prompt, we generate videos using three methods: LayerFlow~\cite{ji2025layerflow}, LayerT2V with Native Mask VAE, and LayerT2V with VAE LoRA
  (Ours).

  \paragraph{Evaluation Protocol.} For each prompt, participants are presented with the three
  generated videos in randomized order (labeled A, B, C) to eliminate ordering bias. They are
  asked to evaluate three aspects: \begin{itemize} \item \textbf{Aesthetic Quality}: Overall
  visual appeal, including color harmony, composition, and absence of artifacts. \item
  \textbf{Foreground Quality}: Completeness, clarity, and temporal stability of the foreground
  layer and its alpha matte. \item \textbf{Text Alignment}: How well the generated video
  content matches the input text prompt. \end{itemize} For each aspect, participants select the
   single best video among the three options. The study interface is shown in
  Fig.~\ref{fig:userstudy}.

  \paragraph{Results.} We compute the preference rate for each method as the fraction of times
  it is selected as the best across all participants and prompts. As reported in Table~\ref{tab:userstudy} of the
   main paper, LayerT2V (Ours) achieves the highest preference rates across all three aspects
  (Aesthetic: 0.724, FG Quality: 0.768, Text Alignment: 0.668), with particularly strong
  performance on Foreground Quality. This confirms that our method produces visually superior
  and more coherent multi-layer videos compared to baselines.

\section{More Visual Samples}
\begin{figure}[h!]
    \centering
    \includegraphics[width=0.98\linewidth]{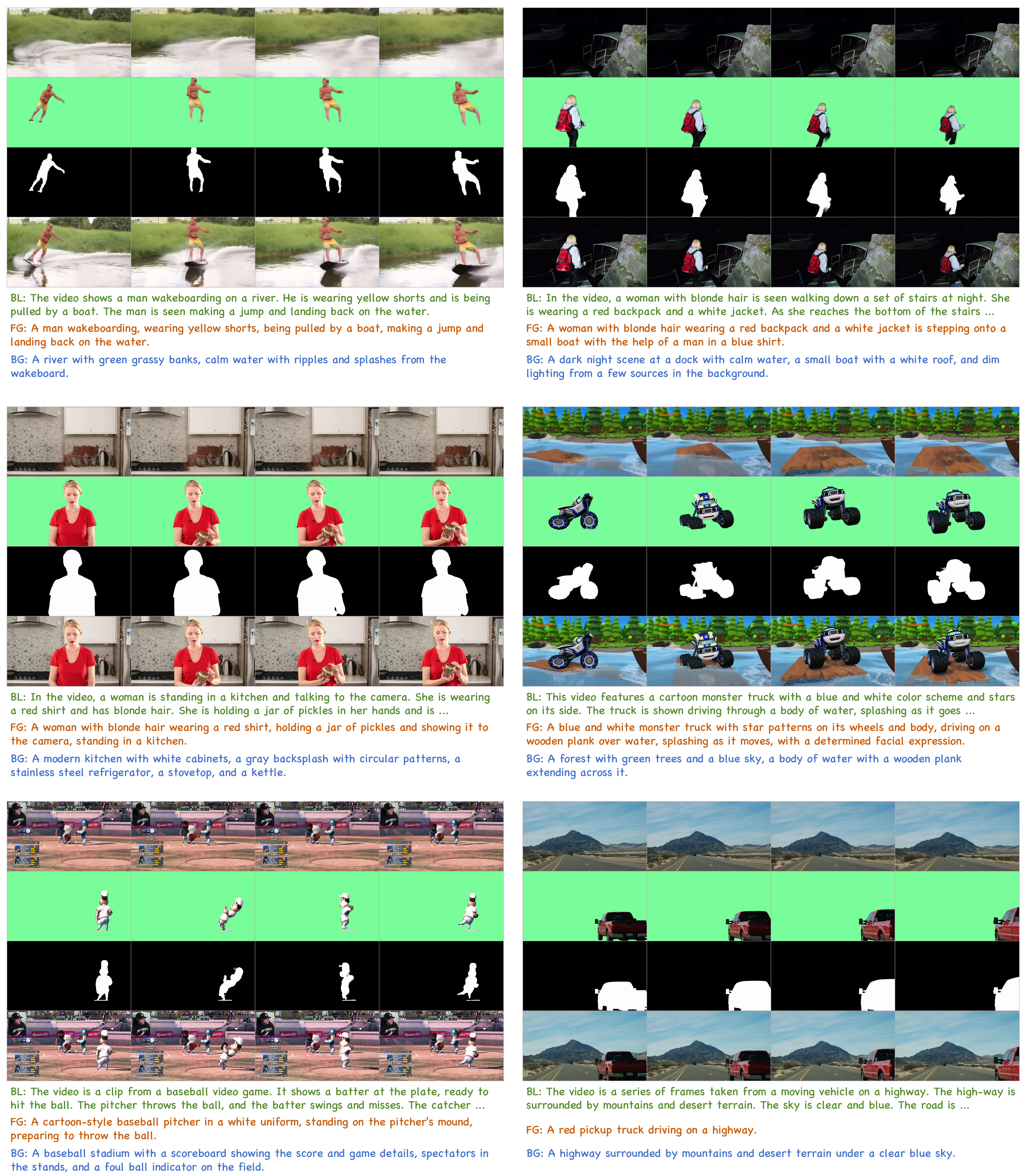}
    \caption{Here, we present additional visualized samples from our proposed VidLayer dataset. The dataset covers a wide and diverse range of scenarios, including human portraits, sports activities, pets, and outdoor environments. It achieves effective and reliable separation of foreground and background content, thereby providing strong and informative supervisory signals for future research on layered content generation, video decomposition, inpainting, and editing tasks.}
    \label{appendix:dataset_sample}
\end{figure}

\begin{figure}[h!]
    \raggedleft
    \includegraphics[width=0.98\linewidth]{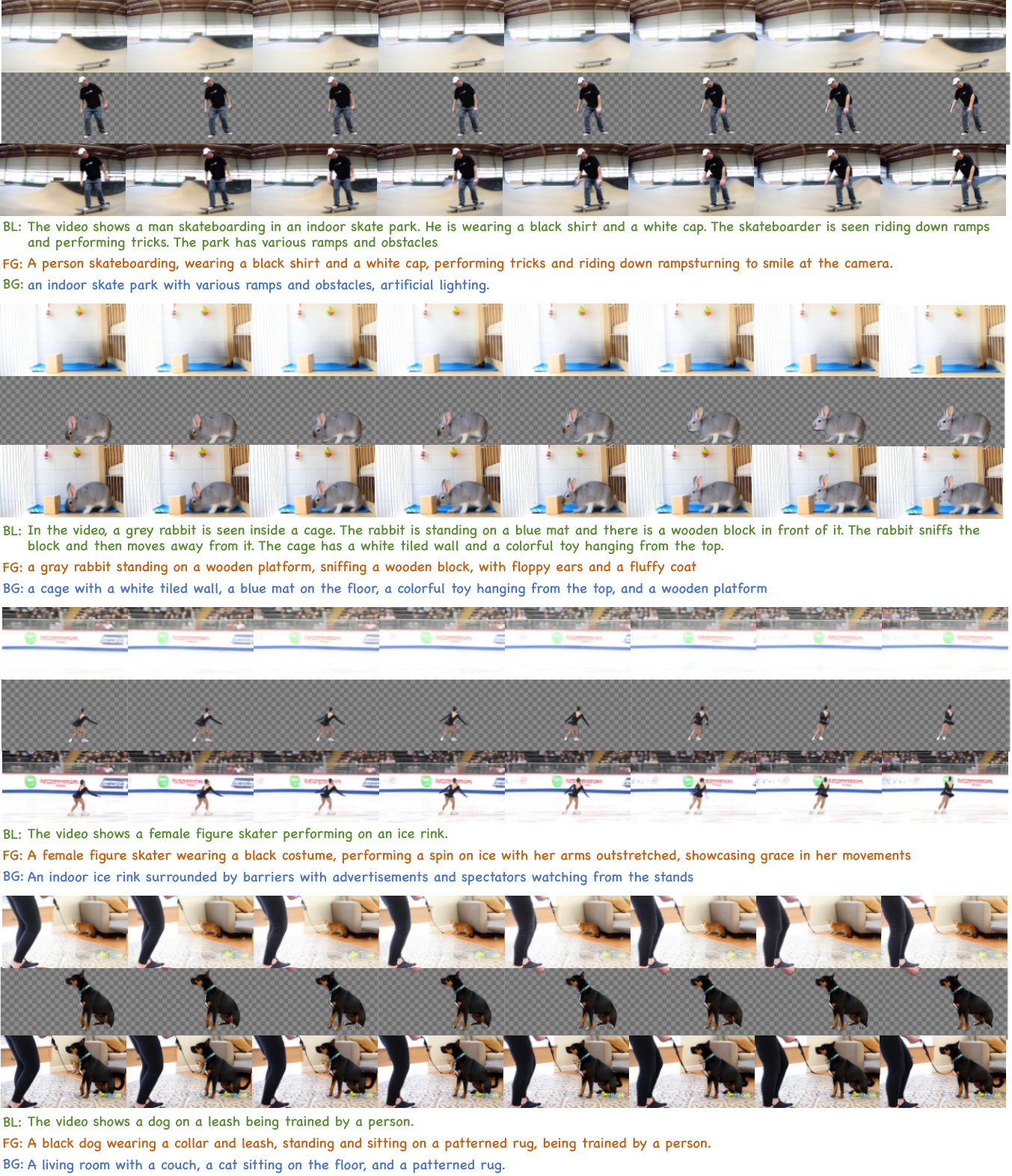}
    \caption{\textbf{Additional qualitative results of LayerT2V.} We show more examples with eight sampled frames per prompt. LayerT2V generates high-fidelity layered outputs, producing clean backgrounds, accurate transparent foregrounds, and coherent recompositions across diverse motions and scenes; the accompanying text provides the full-video description and the layer-specific foreground and background descriptions.}
    \label{appendix:more_result_01}
\end{figure}

\begin{figure}[h!]
    \raggedleft
    \includegraphics[width=0.98\linewidth]{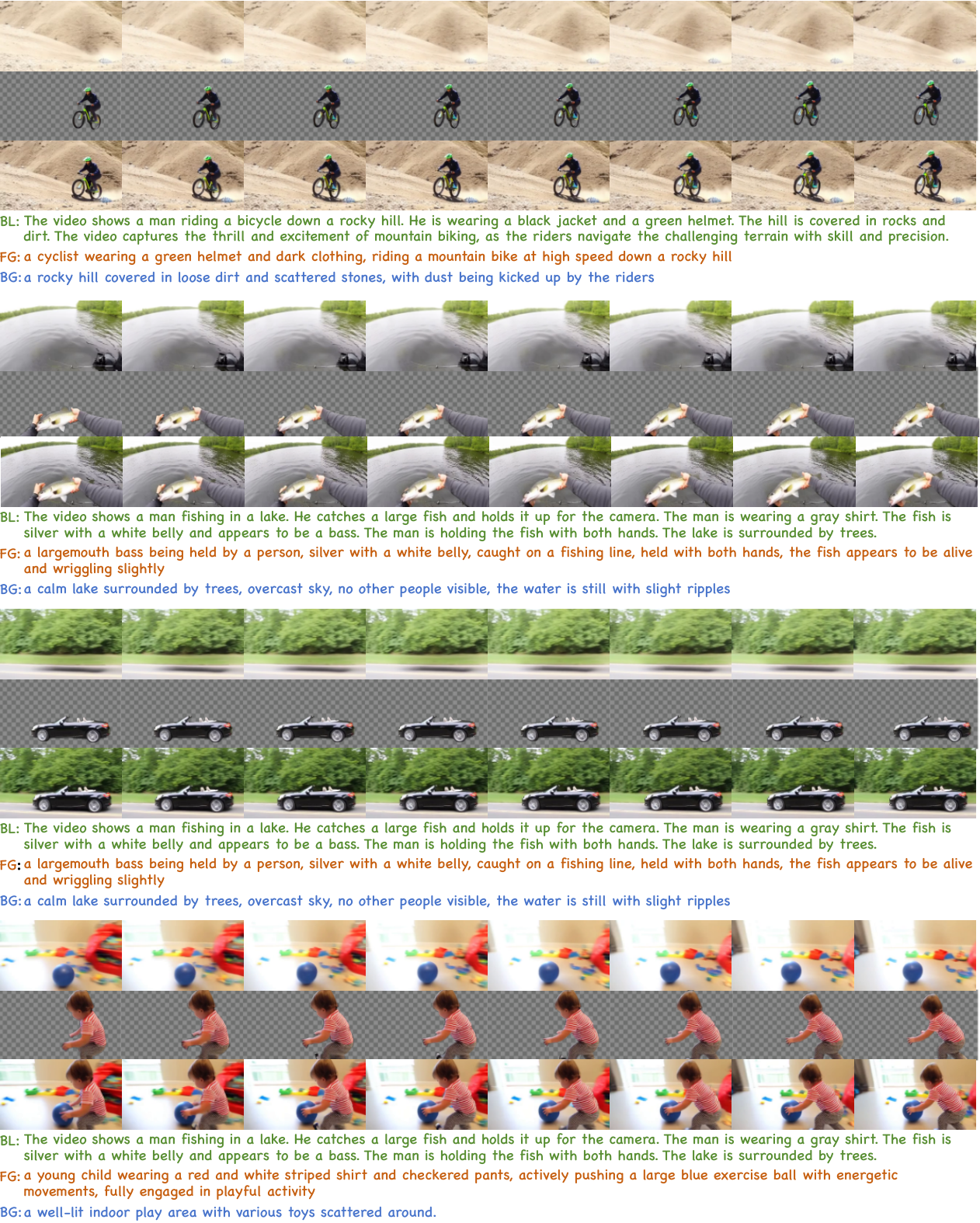}
    \caption{\textbf{Additional qualitative results of LayerT2V.} We show more examples with eight sampled frames per prompt. LayerT2V generates high-fidelity layered outputs, producing clean backgrounds, accurate transparent foregrounds, and coherent recompositions across diverse motions and scenes; the accompanying text provides the full-video description and the layer-specific foreground and background descriptions.}
    \label{appendix:more_result_02}
\end{figure}

\begin{figure}[h!]
    \raggedleft
    \includegraphics[width=0.98\linewidth]{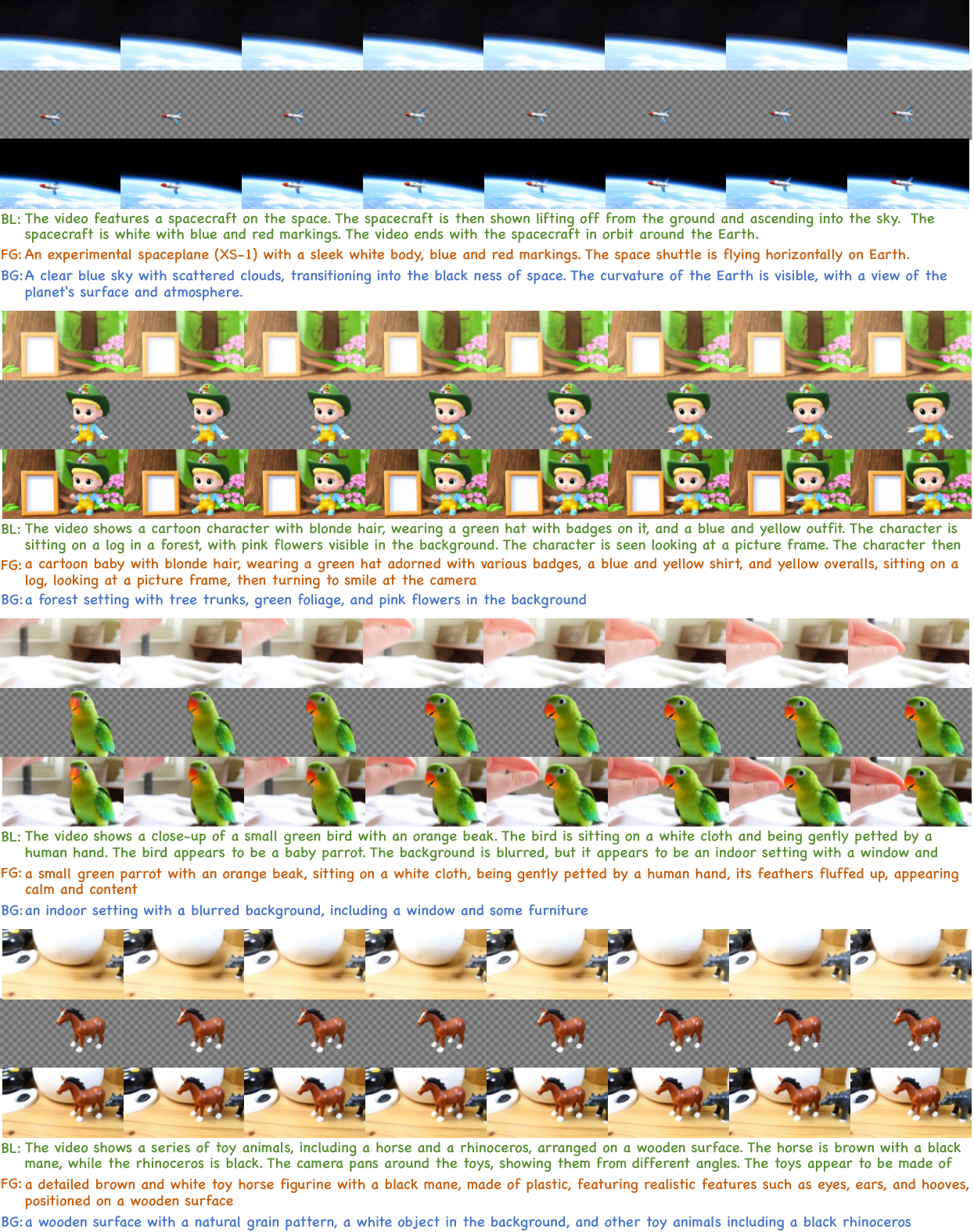}
    \caption{\textbf{Additional qualitative results of LayerT2V.} We show more examples with eight sampled frames per prompt. LayerT2V generates high-fidelity layered outputs, producing clean backgrounds, accurate transparent foregrounds, and coherent recompositions across diverse motions and scenes; the accompanying text provides the full-video description and the layer-specific foreground and background descriptions.}
    \label{appendix:more_result_03}
\end{figure}

\begin{figure}[t!]
    \raggedleft
    \includegraphics[width=0.98\linewidth]{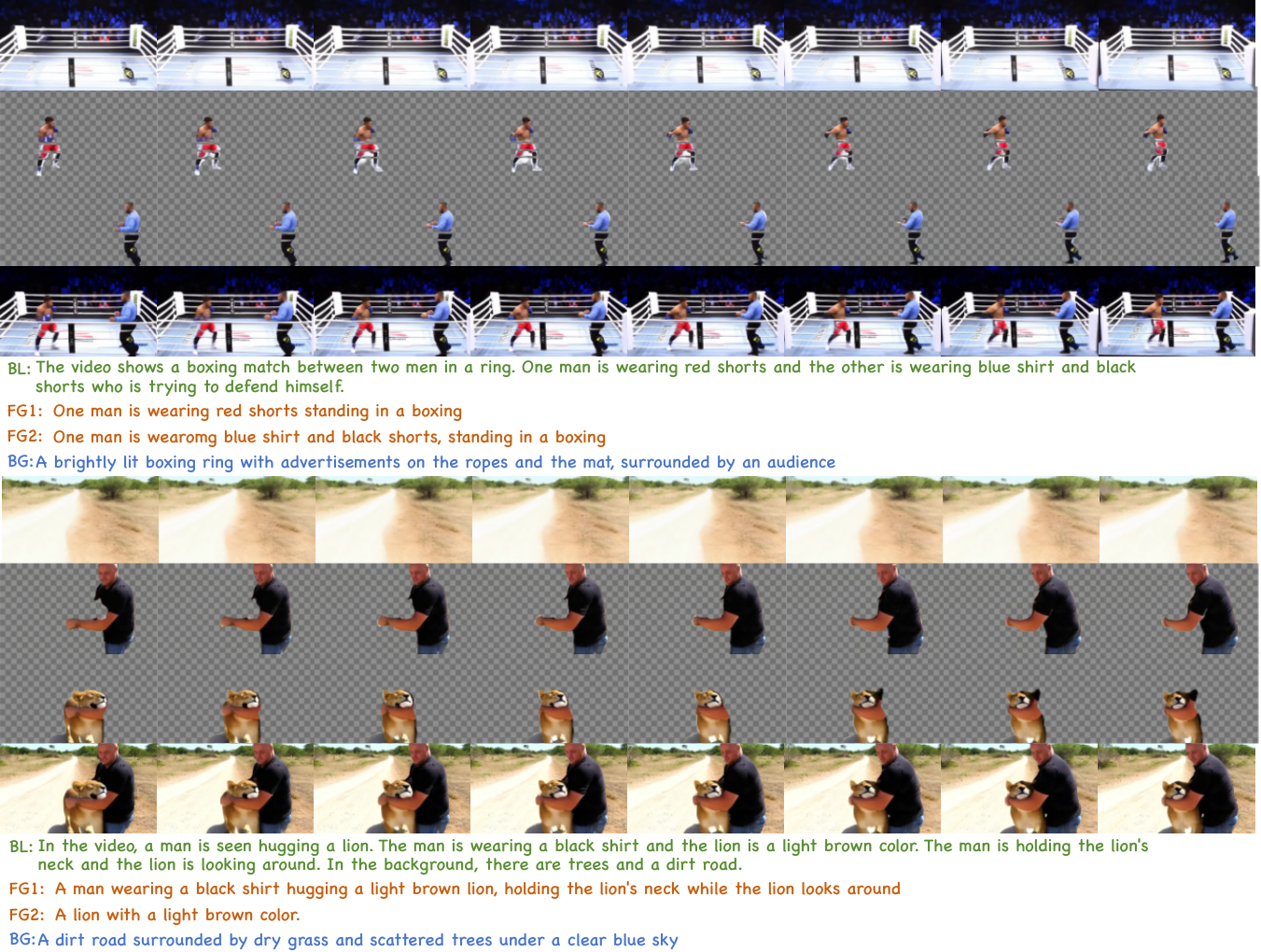}
    \caption{\textbf{Additional two-foreground-layer qualitative results of LayerT2V.} We show more examples with eight sampled frames per prompt. LayerT2V generates high-fidelity layered outputs, producing clean backgrounds, accurate transparent foregrounds, and coherent recompositions across diverse motions and scenes. Extending models to more foreground layers could be a promising future work.}
    \label{appendix:more-res-2-layer}
\end{figure}

In this section, we provide additional visual results to complement the main paper. We first present more samples from our VidLayer dataset (Fig.~\ref{appendix:dataset_sample}). We then include extensive generations of LayerT2V (Figs.~\ref{appendix:more_result_01}--\ref{appendix:more_result_03}), where each example shows eight sampled frames together with the predicted background, transparent foreground, and recomposited video, as well as the full-video prompt and layer-specific descriptions. These results further demonstrate the visual fidelity, temporal stability, and cross-layer coherence of LayerT2V across diverse motions, appearances, and scenes. Lastly, we present additional demonstrations of LayerT2V with two foreground layers in Figure~\ref{appendix:more-res-2-layer}. Owing to computational and resource constraints, our current implementation is limited to modeling two foreground layers. Extending the framework to support a larger number of foreground layers constitutes a promising direction for future work.

\section{Limitations} 
\label{sup:limitations} 
VidLayer is currently limited to at most three foreground layers, with the majority of samples containing only a single foreground. This limitation stems from two factors: (1) the inherent difficulty of collecting source videos that can be cleanly decomposed into multiple semantically distinct and interacting foreground entities, and (2) the limited robustness of current decomposition models when handling complex scenes. 

Specifically, MatAnyone~\cite{yang2025matanyone} struggles to accurately segment small foreground objects over extended temporal contexts, while Gen-Omnimatte~\cite{lee2025generative} is prone to texture leakage and blurry blobs during subject removal in cluttered backgrounds. These failure cases are filtered out by our quality control pipeline, but they reduce the yield of multi-foreground samples. 

Nevertheless, our data engine is fully automated and generalizable—it imposes no assumptions on the underlying decomposition or matting models. As more robust video matting and inpainting methods emerge, our pipeline can be readily applied to construct higher-quality and more diverse layered video datasets.

\newpage

\begin{figure}[h!]
    \centering
    \includegraphics[width=0.98\linewidth]{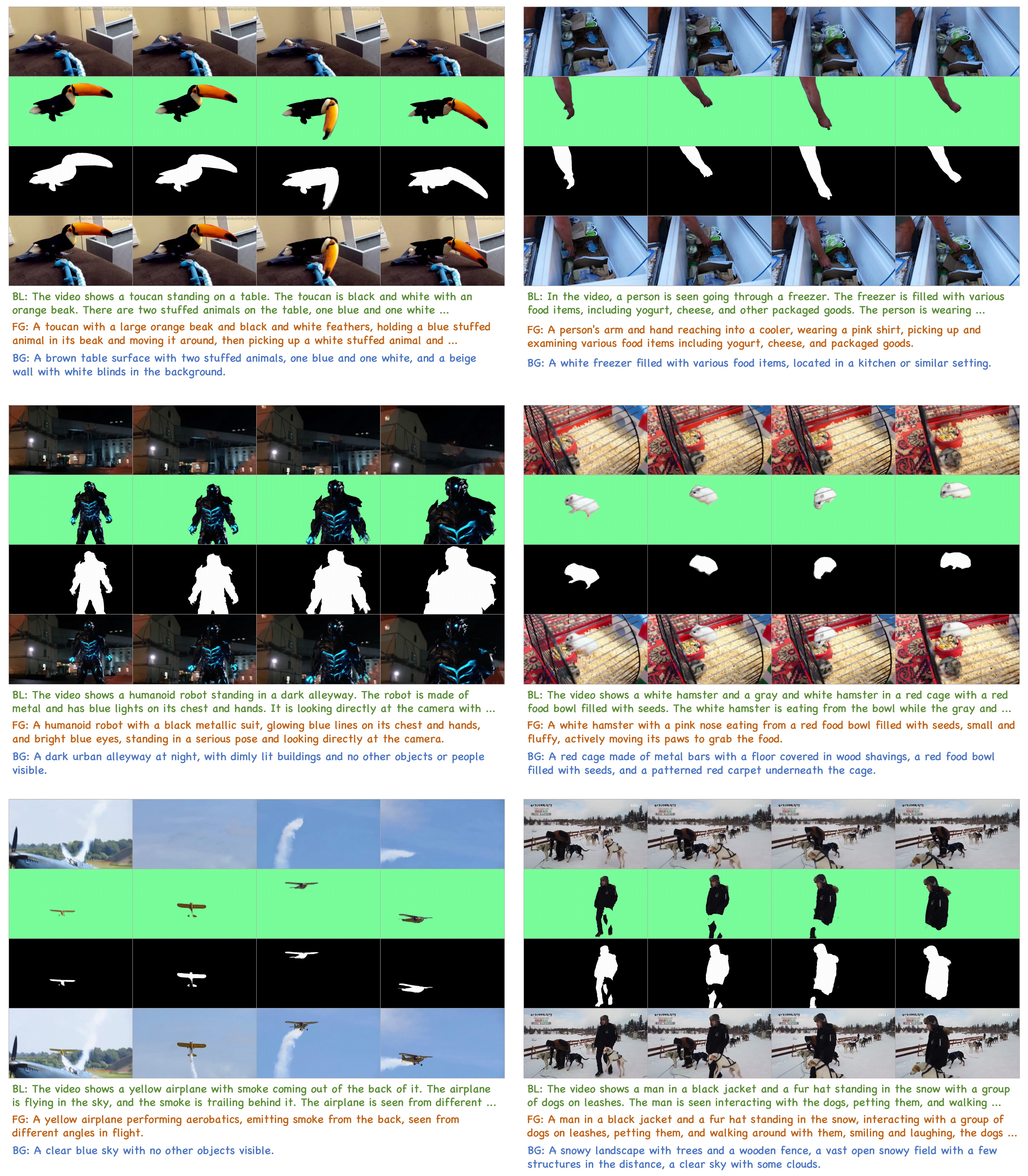}
    \caption{Additional visualized samples from proposed VidLayer dataset.}
    \label{appendix:dataset_sample_2}
\end{figure}

\end{document}